\documentclass[twocolumn]{htl-author}
\usepackage{float}
\usepackage{graphicx}
\usepackage{booktabs} 
\usepackage{caption}
\usepackage{multirow}
\usepackage{tabularx}
\usepackage{xcolor}
\usepackage{makecell}
\usepackage{caption}
\usepackage{longtable}
\definecolor{rebuttal}{rgb}{0, 0, 0}

\DOI{2012.0357}

\usepackage[numbers]{natbib}

\begin{document}

\title{Image Synthesis with Class-Aware Semantic Diffusion Models for Surgical Scene Segmentation}

\author{
  Yihang Zhou$^{1}$, Rebecca Towning$^{2}$, Zaid Awad$^{1,2}$ and Stamatia Giannarou$^{1}$
}

\address{$^{1}$Hamlyn Centre for Robotic Surgery, Department of Surgery and Cancer, Imperial College London, London, SW7 2AZ, UK\\E-mail: yihang.zhou23@imperial.ac.uk\\$^{2}$Imperial College Healthcare NHS Trust, London, W2 1NY, UK
}

\historydate{Published in Healthcare Technology Letters; Received on xxx; Revised on xxx}

\abstract{Surgical scene segmentation is essential for enhancing surgical precision, yet it is frequently compromised by the scarcity and imbalance of available data. To address these challenges, semantic image synthesis methods based on generative adversarial networks and diffusion models have been developed. However, these models often yield non-diverse images and fail to capture small, critical tissue classes, limiting their effectiveness. In response, we propose the Class-Aware Semantic Diffusion Model (CASDM), a novel approach which utilizes segmentation maps as conditions for image synthesis to tackle data scarcity and imbalance. Novel class-aware mean squared error and class-aware self-perceptual loss functions have been defined to prioritize critical, less visible classes, thereby enhancing image quality and relevance. Furthermore, to our knowledge, we are the first to generate multi-class segmentation maps using text prompts in a novel fashion to specify their contents. These maps are then used by CASDM to generate surgical scene images, enhancing datasets for training and validating segmentation models. Our evaluation, which assesses both image quality and downstream segmentation performance, demonstrates the strong effectiveness and generalisability of CASDM in producing realistic image-map pairs, significantly advancing surgical scene segmentation across diverse and challenging datasets.}

\maketitle

\section{Introduction}

Accurate segmentation of surgical scenes is critical for assisting intraoperative navigation and enhancing surgical accuracy, directly influencing surgical outcomes. For example, precise delineation of anatomical structures \cite{madani2022artificial} and accurate detection of tumors and polyps \cite{ali2021deep} can guide the surgeon during the execution of surgical tasks.  Recently developed deep learning models have demonstrated clinically relevant near-real-time segmentation capabilities, some even outperforming human surgeons \cite{kolbinger2023anatomy}.

\textcolor{rebuttal}{Among these approaches, convolutional neural network (CNN)-based methods like DeepLab \cite{chen2017rethinkingatrousconvolutionsemantic} have advanced the field by using atrous convolution to effectively capture detailed features across varying scales, making them adept at detecting both large and subtle structures. U-Net \cite{ronneberger2015unetconvolutionalnetworksbiomedical}, with its encoder-decoder architecture and skip connections, is particularly effective at preserving spatial resolution, which is crucial for the precise segmentation of small anatomical features. More recently, transformer-based models such as SegFormer \cite{xie2021segformersimpleefficientdesign} and Mask2Former \cite{cheng2022maskedattentionmasktransformeruniversal} have emerged, offering significant advantages in capturing global context and multi-scale information through self-attention mechanisms. These transformer-based models have proven particularly valuable in handling complex and diverse surgical scenes, achieving notable improvements in accuracy \cite{kolbinger2023anatomy,psychogyios2024sarrarp50segmentationsurgicalinstrumentation,ayobi2023matis}.}

\textcolor{rebuttal}{However, two major challenges persist across all these segmentation techniques. First, most multi-class segmentation methods are highly data-hungry, requiring large amounts of annotated data for training. Obtaining such data is labor-intensive and time-consuming, and even professional surgeons often struggle to accurately annotate low-contrast areas and unclear edges. Second, while these methods excel at segmenting large and distinct anatomical structures or surgical tools, they frequently struggle with underrepresented classes, particularly in datasets where certain categories are significantly smaller or less prevalent than others. This imbalance can lead to overfitting on these infrequent classes, resulting in poor generalization during testing. Such challenges are particularly critical in surgical applications, where the accurate identification of subtle anomalies can be life-saving.}

To address these issues, generative adversarial networks (GANs) \cite{10.1145/3422622, mahapatra2018efficient} have been proposed for data synthesis, \textcolor{rebuttal}{particularly to augment underrepresented classes.} However, GANs often suffer from mode collapse, producing only a limited variety of data patterns, which limits their effectiveness \cite{thanh2020catastrophic}.

More recently, diffusion models have demonstrated superior abilities in generating high-fidelity and diverse images, setting a new standard in image synthesis \cite{DBLP:journals/corr/abs-2006-11239, DBLP:journals/corr/abs-2010-02502, ozbey2023unsupervised, kazerouni2023diffusion}. Semantic diffusion models have been proposed to synthesize images from segmentation maps, which can then be directly employed in downstream segmentation tasks \cite{wang2022semantic}. For instance, Yuhao et al. developed a framework for colonoscopy image synthesis \cite{du2023arsdm}, while Yan et al. focused on abdominal CT image synthesis \cite{zhuang2023semantic}. However, most diffusion models use the pixel-wise mean squared error (MSE) loss, which inadequately represents smaller classes and leads to their underrepresentation. Moreover, MSE loss does not effectively capture higher-level perceptual features such as textures and structures, often resulting in visually unrealistic images. Additionally, synthetic images generated by semantic diffusion models must adhere to the structure and class distribution predefined by the segmentation maps. This further limits the model's ability to generate diverse variations. 

To overcome the above limitations, we propose the Class-Aware Semantic Diffusion Model (CASDM) tailored for surgical scene generation.
(2) Multi-class segmentation map generation: For the first time, segmentation maps containing multiple classes are generated to guide semantic image synthesis. (3) Text-prompted segmentation map generator: We propose a model that uniquely generates segmentation maps using separate text prompts for each class, specifying class names, quantities, and locations. This method, when integrated with CASDM, yields pairs of high-quality segmentation maps and images, markedly improving dataset diversity. (4) Empirical Validation: Through extensive qualitative and quantitative assessments, our pipeline demonstrates substantial improvements in both image quality and segmentation performance. \textcolor{rebuttal}{We applied our approach to the CholecSeg8K dataset \cite{twinanda2016endonetdeeparchitecturerecognition}, focusing on scarce and challenging classes such as the hepatic vein, blood, and cystic duct. To demonstrate the broader applicability of our model, we also tested it on well-represented classes within the same dataset, confirming that our approach is particularly beneficial for small and rare classes. Additionally, to evaluate the generalisability of our method, we applied it to the gastrectomy SISVSE dataset \cite{yoon2022surgical}. Across different datasets and scenarios, our methodology consistently enhanced segmentation accuracy, underscoring its effectiveness and versatility.}

\begin{figure}[t] 
\centering{\includegraphics[width=8.5cm]{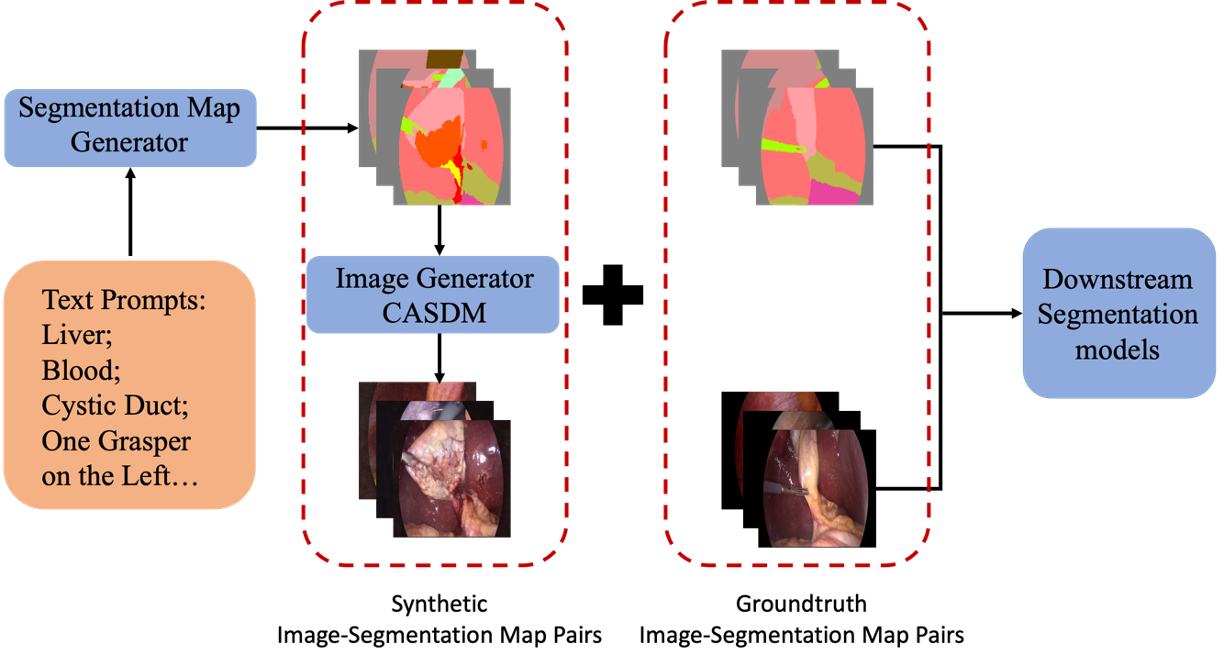}}
\caption{The proposed pipeline, illustrating the process from segmentation map generation through semantic image synthesis to downstream segmentation tasks.}
\label{fig1}
\end{figure}

\section{Method}
Our pipeline, illustrated in Fig. 1, features two primary components: the semantic image generator (CASDM) and the text-guided segmentation map generator. Both components utilize unconditional denoising diffusion probabilistic models (DDPMs)  \cite{DBLP:journals/corr/abs-2006-11239}. 

\emph{\textbf{Unconditional Denoising Diffusion Probabilistic Models (DDPMs):}} 
DDPMs are generative models that synthesize images by progressively adding Gaussian noise in a forward process and subsequently reversing this noise addition in a reverse process, thereby learning to reconstruct data.

In the forward process, the model gradually introduces noise to the data through a Markov chain at each timestep \( t \), transitioning the data \( \mathbf{x}_{t-1} \) to a noisier state \( \mathbf{x}_t \). This process is mathematically described by:
\begin{equation}
q(\mathbf{x}_t | \mathbf{x}_{t-1}) = \mathcal{N}(\mathbf{x}_t; \sqrt{1-\beta_t} \mathbf{x}_{t-1}, \beta_t \mathbf{I})
\end{equation}
where, \( \beta_t \) are the predefined noise schedule parameters incrementally increasing the noise level, and \( \mathcal{N} \) represents the Gaussian distribution. The reverse diffusion process aims to reconstruct the original data by reversing the noise added during the forward process. At each reverse timestep \( t \), the model estimates the less noisy previous state \( \mathbf{x}_{t-1} \) from the current state \( \mathbf{x}_t \) using a conditional distribution parameterized as:

\begin{equation}
p_{\theta}(\mathbf{x}_{t-1} | \mathbf{x}_t) = \mathcal{N}(\mathbf{x}_{t-1}; \mu_{\theta}(\mathbf{x}_t, t), \sigma_t^2 \mathbf{I})
\label{ori_p}
\end{equation}
where \( \mu_{\theta} \) and \( \sigma_t \) are functions of the neural network parameters \(\theta\) (with the network typically being a U-Net architecture), and \(t\), predicting the mean and variance necessary for effective denoising.

The training of DDPM primarily focuses on learning the parameters \( \theta \) that effectively reverse the forward diffusion process. This is achieved by minimizing the pixel-wise MSE between the real noise \( \epsilon \) used in the forward process and the noise predicted by the model \( \hat{\epsilon}_{\theta}(\mathbf{x}_t, t) \) during the reverse process:
\begin{equation}
\mathcal{L}_{\text{MSE}} = \mathbb{E}\left[ \| \epsilon - \hat{\epsilon}_{\theta}(\mathbf{x}_t, t) \|^2 \right]
\label{ori_loss}
\end{equation}
where the expectation is over all data points and all noise levels, optimizing the model to regenerate new data \( \mathbf{x}_0 \) accurately by traversing backward along the Markov chain.

\begin{figure*}[t]
\centering{\includegraphics[width=\textwidth]{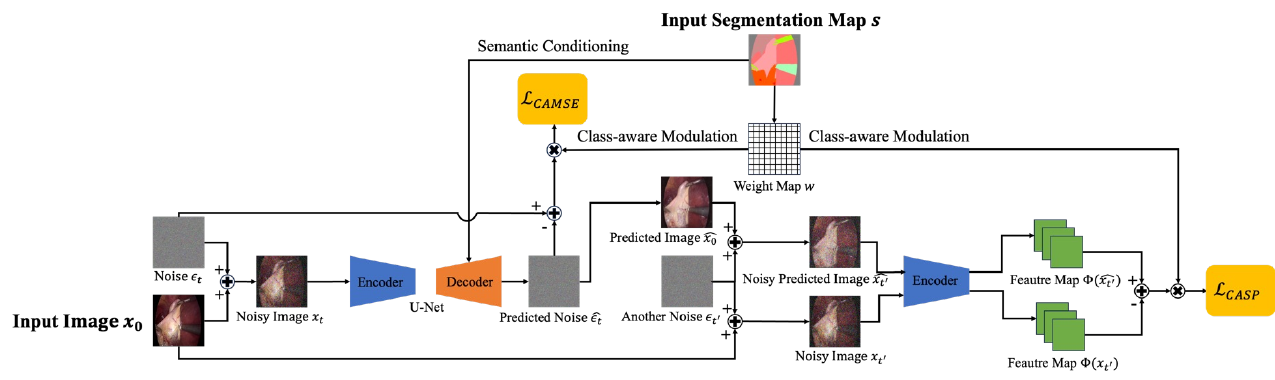}}
\caption{The architecture of CASDM. The inputs are the left image and the top segmentation map. The left side performs semantic image synthesis, calculating the class-aware MSE loss \( \mathcal{L}_{\text{CAMSE}} \). The right side refines the process by comparing the noisy versions of the predicted image and the original input image using the same encoder as the left side, calculating the class-aware self-perceptual loss \( \mathcal{L}_{\text{CASP}} \).}
\label{fig2}
\end{figure*}

\emph{\textbf{Semantic Conditioning:}} The unconditional DDPM does not incorporate semantic conditions into the image generation process, resulting in unpredictable and random images. Additionally, the generated images lack class annotation information, limiting their usefulness in downstream tasks such as segmentation and object detection.
To advance the traditional unconditional DDPMs, we designed the CASDM which is based on semantic image synthesis \cite{wang2022semantic}. CASDM employs a dual-pathway architecture, as shown in Fig. \ref{fig2}. As part of the forward diffusion process, noise is added to the input image which is then processed by the encoder, while the decoder is conditioned by the segmentation map. 
Semantic information is integrated at multiple stages of the decoder using spatially-adaptive normalization layers \cite{wang2022semantic}, ensuring that the generated images maintain their semantic structure and integrity throughout the generation process. This architecture allows CASDM to produce images that have corresponding class annotations (segmentation maps), useful for downstream segmentation tasks.

At the reverse diffusion process, the reverse diffusion equation is based on Equation (\ref{ori_p}), incorporating semantic guidance as:
\begin{equation}
p_{\theta}(\mathbf{x}_{t-1} | \mathbf{x}_t, \mathbf{s}) = \mathcal{N}(\mathbf{x}_{t-1}; \mu_{\theta}(\mathbf{x}_t, \mathbf{s}, t), \sigma_t^2 \mathbf{I})
\end{equation}
where, \( \mathbf{s} \) is the segmentation map. 
The MSE loss from Equation (\ref{ori_loss}) is adapted to:
\begin{equation}
\mathcal{L}_{\text{Semantic MSE}} = \mathbb{E}\left[ \| \epsilon - \hat{\epsilon}_{\theta}(\mathbf{x}_t, \mathbf{s}, t) \|^2 \right]
\end{equation}\

\emph{\textbf{Class-Aware MSE Loss:}} Traditional pixel-wise MSE loss in diffusion models tends to overemphasize larger areas like backgrounds, neglecting small but crucial classes such as blood vessels in surgical scenes. To enhance the fidelity of CASDM to minor classes, we designed the novel class-aware MSE loss function \( \mathcal{L}_{\text{CAMSE}} \). This function ensures all classes, regardless of their size, are accurately represented in the generated images and it is defined as:

\begin{equation}
\mathcal{L}_{\text{CAMSE}} = \sum_{c=1}^{C} w_c \cdot \mathbb{E}\left[ \| \epsilon_c - \hat{\epsilon}_{\theta}(\mathbf{x}_t, \mathbf{s}, t)_c \|^2 \right]
\end{equation}
where, the class weight \( w_c \) is calculated based on the inverse of the pixel count for each class within the segmentation conditioning map, normalized across all classes:
\begin{equation}
w_c = \frac{(H \cdot W / n_c)}{\sum_{j=1}^{C} (H \cdot W / n_j)}
\label{wc}
\end{equation}
where \( H \cdot W \) is the total number of pixels in the image, and \( n_c \) is the number of pixels for class \( c \).

Our approach gives adequate focus to smaller but important tissue classes. This is beneficial for generating pairs of aligned images and segmentation maps that enhance downstream segmentation performance on imbalanced datasets, crucial for surgical imaging applications.\

\emph{\textbf{Class-Aware Self-Perceptual Loss: }}While \( \mathcal{L}_{\text{CAMSE}} \) significantly advances the precision of class representation in semantic image synthesis by ensuring equal attention to all classes, it still primarily focuses on reducing pixel-level discrepancies between the ground truth noise and predicted noise. This function does not capture complex perceptual qualities such as textures and contours, which are essential for producing realistic and clinically applicable images. Moreover, although \( \mathcal{L}_{\text{CAMSE}} \) addresses class imbalance, it might not fully represent the contextual and spatial relationships between classes, which are key for high-fidelity surgical imaging.

Previous approaches have used external segmentation models to refine the training process of semantic image synthesis \cite{du2023arsdm}, but these can introduce additional complexity and potential imperfections. classifier-free guidance (CFG) \cite{ho2022classifier} has also been employed to enhance image quality by interpolating between conditional and unconditional outputs, improving sharpness, fidelity, and diversity. However, CFG can disrupt the coherence between the generated images and the segmentation map conditions. Recent findings \cite{lin2024diffusionmodelperceptualloss} suggest that self-perceptual loss can improve the quality of synthesized images by focusing on high-level perceptual features like textures and structures, thus enhancing visual realism without affecting the constraints of conditions.

To build on these advancements, we introduce the novel class-aware self-perceptual loss \( \mathcal{L}_{\text{CASP}} \) for semantic image synthesis to enrich the perceptual quality of images. As shown in Fig. \ref{fig2} (right refinement side), this loss is computed by first adding identical random noise to both the input image and the image predicted by the diffusion model. These images are then processed through the encoder of the U-Net in the CASDM framework to generate corresponding feature maps from both images. \( \mathcal{L}_{\text{CASP}} \) then uses these feature maps to measure perceptual discrepancies. \( \mathcal{L}_{\text{CASP}} \) specifically amplifies the representation of minor but critical classes by incorporating class-aware considerations into the calculation. It is defined as follows:
\begin{equation}
\mathcal{L}_{\text{CASP}} = \sum_{c=1}^{C} \left\| \left(\Phi(\mathbf{x}_t)_c - \Phi(\hat{\mathbf{x}}_0 + \epsilon)_c\right) \odot \text{pool}(w_c) \right\|^2
\end{equation}
where, \(\Phi\) represents the feature maps extracted by the U-Net encoder, and \(\odot\) signifies element-wise multiplication. The class weights \( w_c \) are derived from Equation (\ref{wc}). The function \( \text{pool}(w_c) \) adapts the class weights to the dimensions of the feature maps.
The predicted image \( \hat{\mathbf{x}}_0 \) is obtained by first predicting the noise output using the core U-Net of the diffusion model. This is given by:
\begin{equation}
\hat{\epsilon}_{\theta} = \epsilon_{\theta}(\mathbf{x}_t, t)
\end{equation}
Using this estimated noise, the original clean image is reconstructed by reversing the noise process applied during the diffusion model's forward process.:
\begin{equation}
\hat{\mathbf{x}}_0 = \frac{\mathbf{x}_t - \sqrt{1 - \beta_t} \hat{\epsilon}_{\theta}}{\sqrt{\alpha_t}}
\end{equation}
where, \( \alpha_t = \prod_{i=1}^{t} (1 - \beta_i) \) and \( \beta_t \) is the noise schedule.

The advantages of using this novel \( \mathcal{L}_{\text{CASP}} \) is that we avoid the use of additional modulation models, ensuring that the generated images align with the conditions while enhancing image details related to textures, colors, and structures, especially for smaller classes. 

\begin{figure}[t]\label{fig3}
\centering{\includegraphics[width=8.5cm]{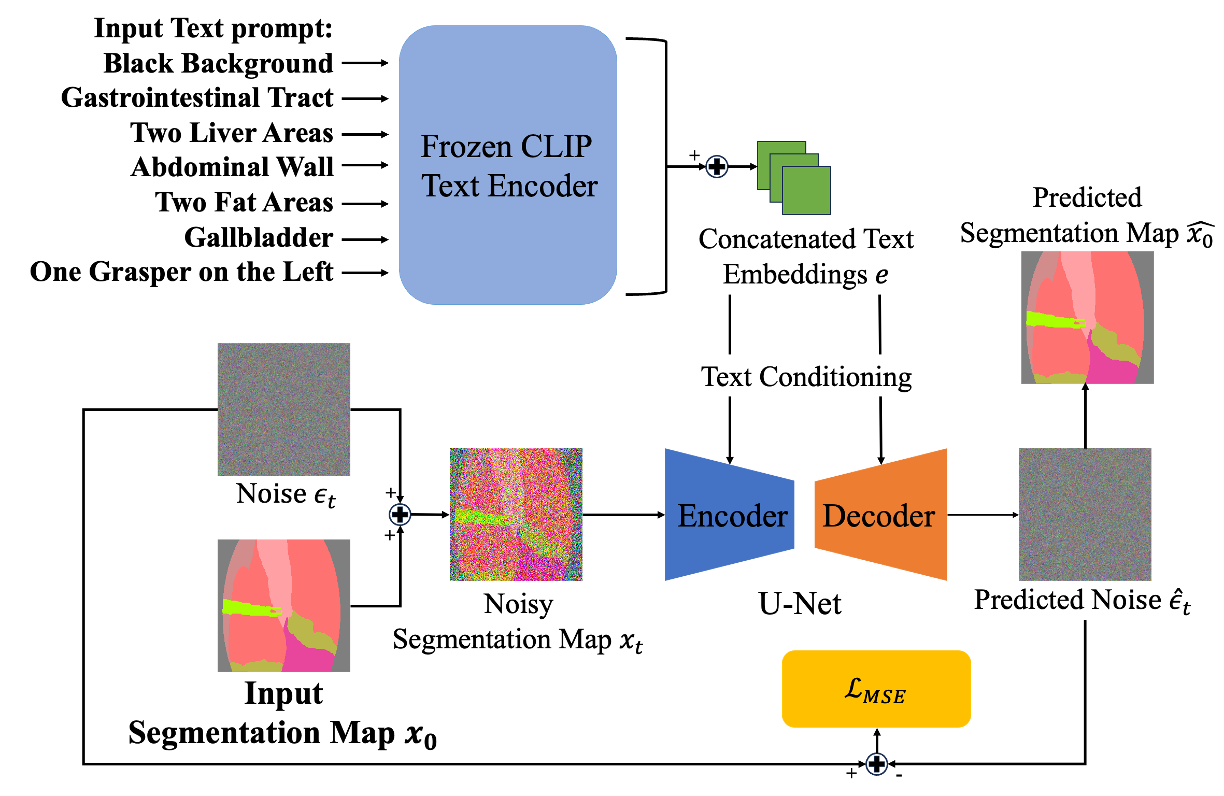}}
\caption{The architecture of the text-prompted segmentation map generator. The inputs are the left segmentation map and separate text prompts specifying class names, quantities, and locations. The output is the pixel-wise MSE loss \( \mathcal{L}_{\text{MSE}} \) on the bottom.}
\end{figure}

\emph{\textbf{Text-Prompted Segmentation Map Generator: }}Using ground truth segmentation maps to generate images inherently limits diversity. To address this issue, we are the first to propose a novel pipeline that uses text prompts in a unique way to control the synthesis of multi-class segmentation maps. These maps are then used in CASDM to create diverse pairs of images and segmentation maps.

The process starts by inputting class names separately for each class into a frozen text encoder based on the contrastive language-image pre-training (CLIP) model \cite{radford2021learningtransferablevisualmodels}, with the option to include directional and quantitative information for each class. This innovative method of separate text input for each class allows for the isolated encoding of class characteristics, significantly enhancing the specificity and relevance of the generated text embeddings. These individual embeddings are then concatenated to form a comprehensive control vector. The CLIP model, pretrained on 400 million image-text pairs, effectively captures detailed visual concepts including locations and quantities from raw text.

The integrated text embeddings modify the conditional distribution in the reverse diffusion steps of this text-guided DDPM:
\begin{equation}
p_{\theta}(\mathbf{x}_{t-1} | \mathbf{x}_t, \mathbf{e}) = \mathcal{N}(\mathbf{x}_{t-1}; \mu_{\theta}(\mathbf{x}_t, \mathbf{e}, t), \sigma_t^2 \mathbf{I})
\end{equation}
where, \(\mathbf{e}\) represents the concatenated text embedding. 

By leveraging text prompts in this manner, we can produce highly diverse and controlled segmentation maps. These maps allow the CASDM to generate corresponding synthetic images, significantly enhancing the dataset's diversity. Our proposed segmentation map generator is illustrated in Fig. 3.

\section{Experiments and Results}

Our experiments aim to show that the proposed framework is able to alleviate the poor segmentation performance caused by data scarcity and class imbalance. We demonstrate that CASDM and our text-prompted segmentation map generation pipeline can produce high-quality synthetic data for underrepresented classes, \textcolor{rebuttal}{and that our approach generalizes well across different datasets. The code and synthetic data for CASDM are available at \url{https://github.com/YihangZhou123/CASDM-Code-and-Synthetic-Data}.}

\emph{\textbf{Dataset: }}We \textcolor{rebuttal}{initially} conducted our experiments on the CholecSeg8K dataset \cite{twinanda2016endonetdeeparchitecturerecognition}. It is specifically curated for surgical semantic segmentation, aiming to enhance the accuracy and safety of computer-assisted cholecystectomy procedures. The 8080 image-mask pairs cover 13 classes, including black background, abdominal wall, liver, gastrointestinal tract, fat, grasper, connective tissue, blood, cystic duct, L-hook electrocautery, gallbladder, hepatic vein, and liver ligament. Three of these classes namely blood, cystic duct, and hepatic vein are underrepresented as they are included only in 8.56\%, 3.07\%, and 3.92\% of the images, respectively. On the images where they are present, these classes also occupy a very small proportion of the total pixel count, with blood at 4.93\%, cystic duct at 1.2\%, and hepatic vein at 0.24\%. Additionally, blood is inherently difficult to distinguish perfectly from surrounding tissues, making it a persistent challenge in segmentation. \textcolor{rebuttal}{To further validate our approach, we repeated our experiments on three other classes to assess its broader applicability. This included the rare but large liver ligament class, present in only 2.97\% of images but occupying an average of 14.1\% of the area in those images, with consistent shape, location, and appearance. The not rare but small L-hook electrocautery class appears in 27.9\% of images, covering an average area of 4.7\%—similar to that of blood. The well-represented connective tissue class is present in 19.8\% of images and occupies an average area of 11.45\%.}

To ensure a robust evaluation of our approach, we split the dataset into 6868 images for training and 1212 images for testing at a ratio of 85:15.




\emph{\textbf{Model Implementation: }}We implemented our CASDM using the PyTorch library and trained it on an NVIDIA RTX A6000 GPU. All image-map pairs from the CholecSeg8K were resized to 256x256, with data augmentation including random horizontal (p=0.5) and vertical (p=0.5) flips, and rotations (p=0.5). The CASDM was trained with a batch size of 10, a learning rate of 1e-4, for approximately 100 epochs, using a traditional DDPM noise scheduler. During inference, we used the DPM-Solver++ noise scheduler \cite{lu2023dpmsolverfastsolverguided} with 30 inference timesteps, resulting in an image-map generation time of approximately 2 seconds. For the segmentation map generator, we applied the same settings as the CASDM, except the input consisted of pairs of segmentation maps and corresponding text prompts. Additionally, we implemented a color mapping technique to ensure that the colors of all generated masks align with those of the target classes. This was achieved by mapping each color to its nearest counterpart in the target color space.\

\emph{\textbf{Comparison: }}To validate our framework, we evaluated the quality of the synthesised images as well as the accuracy of the downstream segmentation task. 

\begin{table}[H]
\centering 
\caption{\textcolor{rebuttal}{Evaluation of the quality of images synthesised by three\\image synthesis models.}}
{ 
\begin{tabular}{|l|c|c|c|}
  \hline
  \textbf{Method} & \textbf{FID $\downarrow$} & \textbf{Mean MS-SSIM $\uparrow$} & \textbf{Mean PSNR $\uparrow$} \\\hline
  \makecell[l]{Conditional \\ DDPM} & 125.16 & 0.62 & 12.00 \\\hline
  SDM & 85.51 & 0.60 & 13.08 \\\hline
  CASDM & \textbf{75.89} & \textbf{0.64} & \textbf{14.44} \\\hline
\end{tabular}
}
\label{table1}
\end{table}

\begin{table*}[t]
\centering
\caption{Comparisons of data expansion methods on segmentation performance \textcolor{rebuttal}{ for underrepresented classes in the CholecSeg8K dataset}.}
\resizebox{\textwidth}{!} {
\begin{tabular}{|l|l|c|c|c|c|c|c|c|c|}
  \hline
  \multirow{2}{*}{\textbf{Seg. Model}} & \multirow{2}{*}{\textbf{Training Dataset}} & \multicolumn{2}{c|}{\textbf{Blood}} & \multicolumn{2}{c|}{\textbf{Cystic Duct}} & \multicolumn{2}{c|}{\textbf{Hepatic Vein}} & \multicolumn{2}{c|}{\textbf{All}} \\\cline{3-10}
  & & \textbf{mIoU $\uparrow$} & \textbf{mDice $\uparrow$} & \textbf{mIoU $\uparrow$} & \textbf{mDice $\uparrow$} & \textbf{mIoU $\uparrow$} & \textbf{mDice $\uparrow$} & \textbf{mIoU $\uparrow$} & \textbf{mDice $\uparrow$} \\\hline
  \multirow{6}{*}{SegFormer} & Original & 79.5 & 88.6 & 81.8 & 90.0 & 65.5 & 79.1 & 89.4 & 94.2 \\\cline{2-10}
  & + Naive Data Augmentation & 78.4 & 87.9 & 80.8 & 89.4 & 66.1 & 79.6 & 89.1 & 94.0 \\\cline{2-10}
  & + Conditional DDPM & 80.2 & 89.0 & 82.5 & 90.4 & 66.9 & 80.2 & 89.6 & 94.3 \\\cline{2-10}
  & + SDM & 80.1 & 88.9 & 82.7 & 90.6 & 66.9 & 80.2 & 89.6 & 94.3 \\\cline{2-10}
  & + CASDM & 80.4 & 89.1 & 83.3 & 90.9 & 67.2 & 79.5 & 89.6 & 94.3 \\\cline{2-10}
  & + CASDM + Synthetic Maps & \textbf{81.5} & \textbf{89.8} & \textbf{83.8} & \textbf{91.2} & 67.8 & 80.8 & 89.9 & \textbf{94.5} \\\hline
  \multirow{6}{*}{Mask2Former} & Original & 78.9 & 88.2 & 80.2 & 89.0 & 66.7 & 80.0 & 89.3 & 94.1 \\\cline{2-10}
  & + Naive Data Augmentation & 78.6 & 88.0 & 81.2 & 89.6 & 65.0 & 78.8 & 89.2 & 94.1 \\\cline{2-10}
  & + Conditional DDPM & 79.2 & 88.4 & 81.7 & 89.9 & 67.8 & 80.8 & 89.7 & 94.4 \\\cline{2-10}
  & + SDM & 79.2 & 88.4 & 81.1 & 89.6 & 67.6 & 80.7 & 89.6 & 94.3 \\\cline{2-10}
  & + CASDM & 79.8 & 88.8 & 82.2 & 90.2 & 67.8 & 80.8 & 89.8 & 94.4 \\\cline{2-10}
  & + CASDM + Synthetic Maps & 80.9 & 89.4 & 83.4 & 90.9 & \textbf{68.5} & \textbf{81.3} & \textbf{90.0} & \textbf{94.5} \\\hline
\end{tabular}
}
\label{table2}
\end{table*}

\begin{table*}[h]
\centering
\caption{\textcolor{rebuttal}{Comparisons of data expansion methods on segmentation performance for well-represented classes in the CholecSeg8K dataset.}}
\textcolor{rebuttal}{\resizebox{\textwidth}{!} {
\begin{tabular}{|l|l|c|c|c|c|c|c|c|c|}
  \hline
  \multirow{2}{*}{\textbf{Seg. Model}} & \multirow{2}{*}{\textbf{Training Dataset}} & \multicolumn{2}{c|}{\textbf{Liver Ligament}} & \multicolumn{2}{c|}{\textbf{Connective Tissue}} & \multicolumn{2}{c|}{\textbf{L-hook Electrocautery}} & \multicolumn{2}{c|}{\textbf{All}} \\\cline{3-10}
  & & \textbf{mIoU $\uparrow$} & \textbf{mDice $\uparrow$} & \textbf{mIoU $\uparrow$} & \textbf{mDice $\uparrow$} & \textbf{mIoU $\uparrow$} & \textbf{mDice $\uparrow$} & \textbf{mIoU $\uparrow$} & \textbf{mDice $\uparrow$} \\\hline
  \multirow{7}{*}{SegFormer} & Original & 96.7 & 98.3 & 90.9 & 95.3 & 91.3 & 95.4 & 89.4 & \textbf{94.2} \\\cline{2-10}
  & + Naive Data Augmentation & 96.8 & 98.4 & 90.9 & 95.2 & 91.4 & 95.5 & 89.4 & \textbf{94.2} \\\cline{2-10}
  & + Conditional DDPM & 96.9 & 98.4 & 90.9 & 95.3 & 91.3 & 95.4 & 89.4 & \textbf{94.2} \\\cline{2-10}
  & + SDM & 97.1 & 98.5 & \textbf{91.1} & \textbf{95.4} & 90.9 & 95.2 & 89.3 & 94.1 \\\cline{2-10}
  & + CASDM & \textbf{97.6} & \textbf{98.8} & 91.0 & 95.3 & 91.7 & 95.7 & \textbf{89.5} & \textbf{94.2} \\\cline{2-10}
  & + CASDM + Synthetic Maps & 96.9 & 98.4 & \textbf{91.1} & \textbf{95.4} & 92.5 & 96.1 & \textbf{89.5} & \textbf{94.2} \\\hline
  \multirow{7}{*}{Mask2Former} & Original & 96.7 & 98.3 & 90.5 & 95.0 & 91.9 & 95.8 & 89.3 & 94.1 \\\cline{2-10}
  & + Naive Data Augmentation & 96.7 & 98.3 & 91.0 & 95.3 & 91.6 & 95.6 & 89.4 & \textbf{94.2} \\\cline{2-10}
  & + Conditional DDPM & 96.6 & 98.3 & 90.9 & 95.3 & 92.5 & 96.1 & 89.4 & \textbf{94.2} \\\cline{2-10}
  & + SDM & 97.1 & 98.5 & 90.8 & 95.2 & 91.4 & 95.5 & \textbf{89.5} & \textbf{94.2} \\\cline{2-10}
  & + CASDM & 97.2 & 98.6 & 90.8 & 95.2 & \textbf{92.7} & \textbf{96.2} & \textbf{89.5} & \textbf{94.2} \\\cline{2-10}
  & + CASDM + Synthetic Maps & 97.3 & 98.6 & \textbf{91.1} & 95.3 & 92.0 & 95.9 & \textbf{89.5} & \textbf{94.2} \\\hline
\end{tabular}}
}
\label{table3}
\end{table*}

For the image quality check, we compared our framework 
with the baseline semantic diffusion model (SDM) \cite{wang2022semantic}, and the conditional DDPM \cite{zhuang2023semantic} which concatenates the input image and the corresponding segmentation map to be used as one input to control the image synthesis. We trained the above models on the original CholecSeg8K training dataset and used all of the segmentation maps from the CholecSeg8K testing dataset to synthesise images. To evaluate the quality of the synthesised images, we use the \textcolor{rebuttal}{Fréchet inception distance (FID)}, the mean multi-scale structural similarity index measure (Mean MS-SSIM) and the mean peak signal-to-noise ratio (Mean PSNR). \textcolor{rebuttal}{FID has been commonly used to quantify the realism and diversity of synthetic images. According to \cite{woodland2024featureextractiongenerativemedical}, FID is a valuable metric to compare models for surgical image synthesis rather than using its absolute value to assess a model's performance.} 
As shown in Table \ref{table1}, CASDM achieves the \textcolor{rebuttal}{best} performance, verifying its ability to synthesize high-quality images adherent to segmentation maps.

\begin{figure}[!h]
\centering{\includegraphics[width=8.5cm]{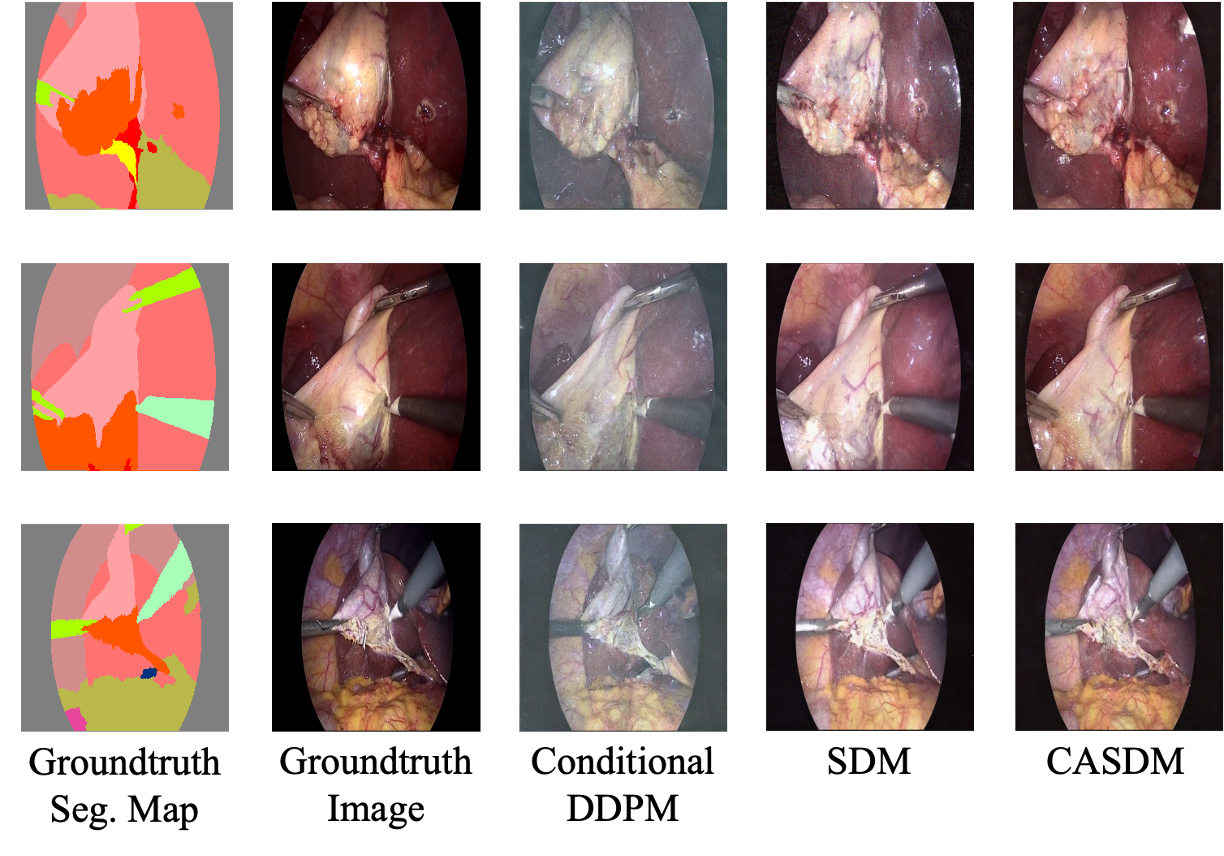}}
\caption{Synthetic images generated by compared image synthesis models.}
\label{fig4}
\end{figure}

As illustrated in Fig. \ref{fig4}, the conditional DDPM demonstrates strong capabilities in synthesising images aligned with segmentation maps. However, it exhibits significant blurriness when the DPM-Solver++ scheduler is used with 30 timesteps. Using the DPM-Solver++ noise scheduler slightly affects image quality, as described in \cite{lu2023dpmsolverfastsolverguided}. The original DDPM noise scheduler with 1000 timesteps can generate more aligned and refined images. However, it would take over 1 minute per image-map pair, making it impractical. SDM produces images that appear satisfactory, but CASDM notably presents more realistic colors. This is particularly impressive given that CASDM, aimed at practical rapid generation, requires only 2 seconds to generate one pair of image and segmentation map.


Next, we conducted a downstream segmentation evaluation using the SegFormer \cite{xie2021segformersimpleefficientdesign} and the Mask2Former \cite{cheng2022maskedattentionmasktransformeruniversal}, which are widely recognized and accepted segmentation models in surgical applications. Both models underwent training from scratch on the original CholecSeg8K training dataset and were tested on the CholecSeg8K testing dataset. 

We then employed the naive data augmentation method that creates several identical copies of data in the dataset, the conditional DDPM, SDM, and CASDM to synthesise images from the segmentation maps of the CholecSeg8K training dataset, which were used to augment the original training data.

To address the class imbalance and to demonstrate the models' capabilities with smaller classes, we specifically selected segmentation maps from the training data containing any of the blood, cystic duct, or hepatic vein classes. For each method, we produced an additional 6,868 images containing these specific classes, significantly enhancing their representation in the training data.
 
Subsequently, our novel text-prompted segmentation map generator was employed to produce synthetic segmentation maps. By ensuring the text prompts matched the patterns found in the existing training set, this approach maintained consistency in the training of the downstream models and prevented excessive variability.
These segmentation maps were then used with CASDM to synthesise corresponding images, doubling the number of training data.

\textcolor{rebuttal}{We repeated the above steps for liver ligament, connective tissue, and L-hook electrocautery—classes that are well-represented in the dataset. This was done to evaluate how our model performs on non-target classes that are not scarce and small, thereby assessing its potential capabilities beyond the primary focus on balancing imbalanced datasets.}

\begin{figure}[!h]
\centering{\includegraphics[width=8.5cm]{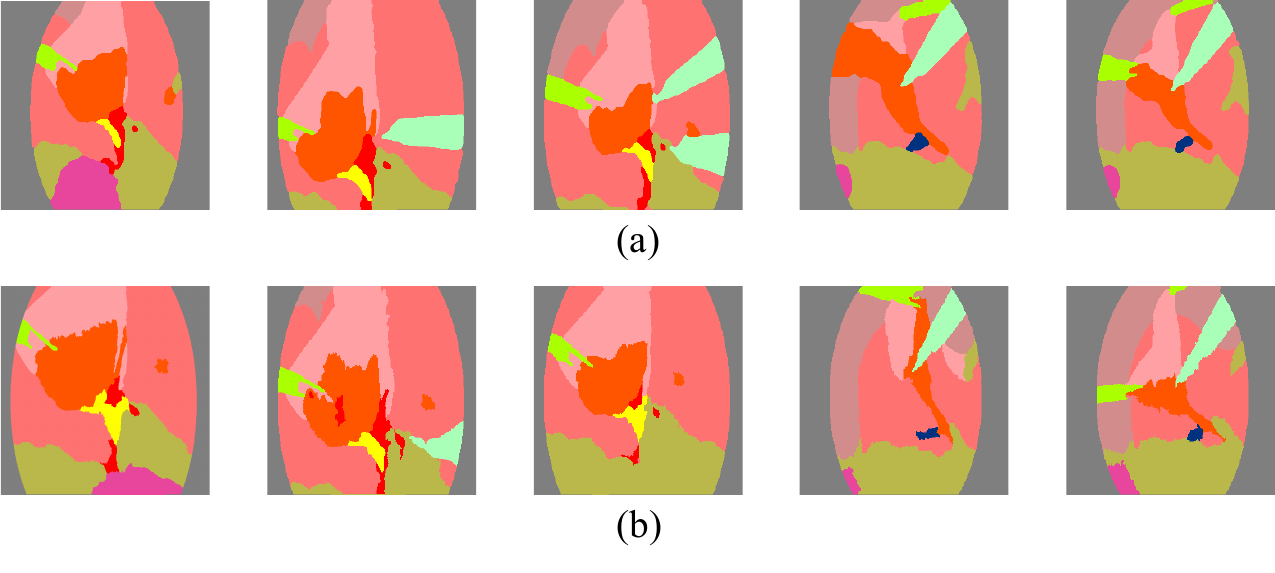}}
\caption{Samples of (a) segmentation maps generated by our text-prompted method and (b) segmentation maps included in the CholecSeg8K dataset.}
\label{fig5}
\end{figure}

As shown in Fig. \ref{fig5}, it is evident that the generated segmentation maps closely match the patterns of the ground truth segmentation maps. Since metrics like \textcolor{rebuttal}{FID measures how visually realistic images are,} Mean MS-SSIM is designed for image reconstruction and Mean PSNR would be artificially high after color mapping for synthesized segmentation maps, they are not suitable for evaluating the quality of segmentation map generation. The assessment of the generated segmentation maps should be based on their effectiveness in enhancing downstream segmentation models when used in conjunction with CASDM to generate new data.

\textcolor{rebuttal}{For evaluating the effectiveness of different data expansion methods on underrepresented classes,} we assessed the segmentation results in terms of the mean intersection over union (mIoU) and mean dice coefficients (mDice) for blood, cystic duct, and hepatic vein on the testing set. As shown in Table \ref{table2}, the naive data augmentation method surpasses the original dataset's performance only for the hepatic vein class with the SegFormer model, and for the cystic duct class with the Mask2Former model. In the other cases, it shows a deterioration in the performance. This is expected as increasing the sample size without enhancing diversity leads to a significant risk of overfitting. When the training data is augmented with synthesised images generated by segmentation maps only from the training set, 
CASDM achieves the highest mIoU and mDice, demonstrating its strict adherence to segmentation map guidance to synthesise meaningful images. CASDM showed an average improvement across these three classes of 1.4\% in mIoU and 0.6\% in mDice for SegFormer, and 1.3\% in mIoU and 0.9\% in mDice for Mask2Former. Additionally, the combination of CASDM with our text-prompted segmentation map generator (CASDM + Synthetic Maps) achieved the highest mIoU and mDice on both SegFormer and Mask2Former, demonstrating the practical utility of the proposed pipeline. On SegFormer, this combination improved the mIoU by 1.8\% and mDice by 1.4\% across the three classes. On Mask2Former, it improved the mIoU by 2\% and mDice by 1.5\%. Furthermore, on Mask2Former, it enhanced the mIoU for the cystic duct by 3.2\%, significantly proving the effectiveness of this approach.

\captionsetup[figure]{labelfont={color=rebuttal,bf},textfont={color=rebuttal}}

\begin{figure*}[t]
\centering{\includegraphics[width=\textwidth]{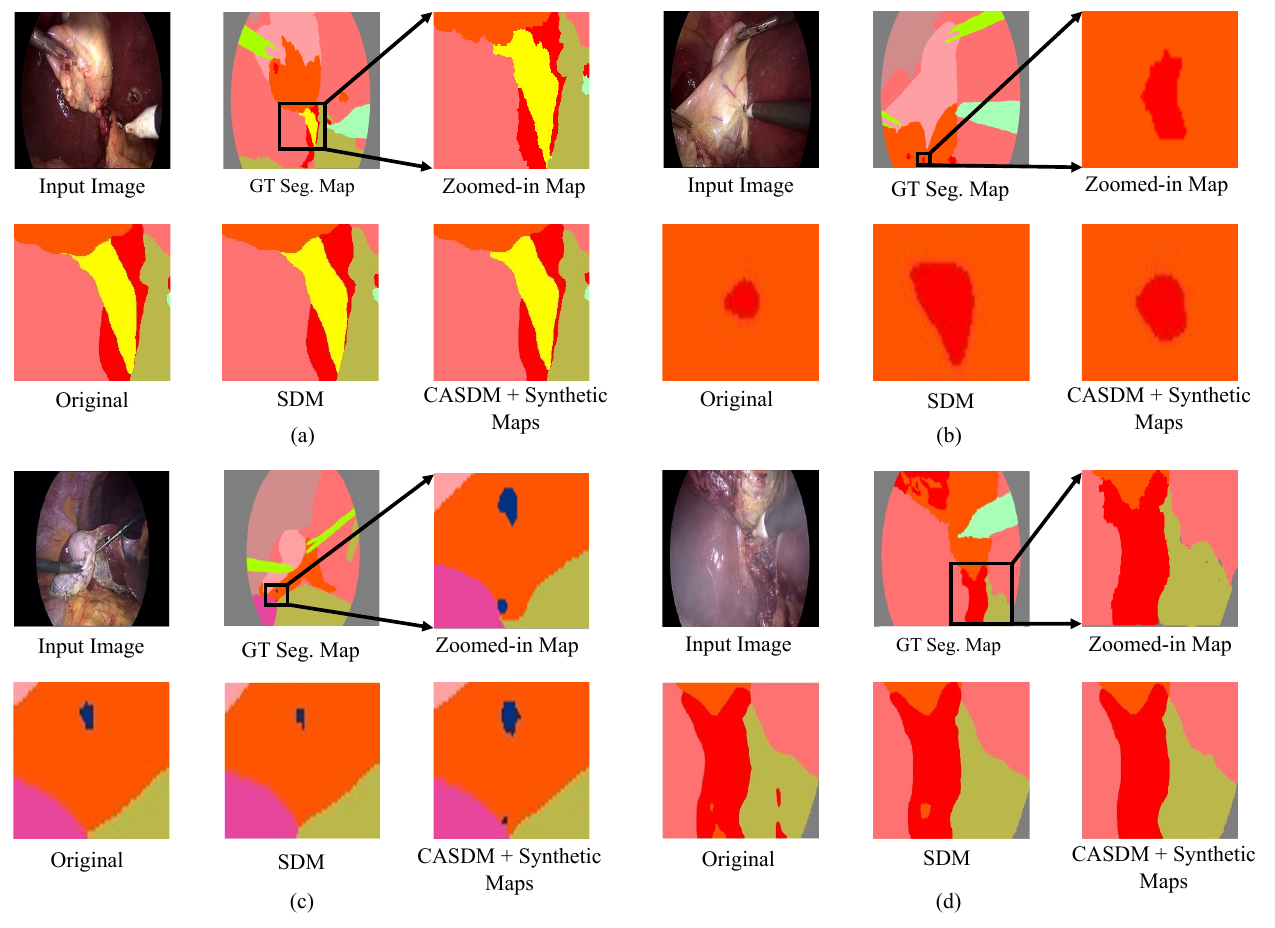}}
\caption{Visual comparison of segmentation models on underrepresented classes. (a) Segmentation of the cystic duct (yellow) with SegFormer. (b) Segmentation of a blood clot (red) with Mask2Former. (c) Segmentation of the hepatic vein (blue) with SegFormer in a slightly smoke-challenged case. (d) Segmentation of the blood (red) with Mask2Former in a highly smoke-challenged case. Models were trained on the original CholecSeg8K dataset, the dataset augmented with SDM-synthesized data, and the dataset augmented with CASDM + synthetic maps.}
\label{fig6}
\end{figure*}
\captionsetup[figure]{labelfont=bf, textfont=normalcolor}

As shown in Fig. \ref{fig6}, segmentation models trained on data augmented by CASDM and synthetic maps perform significantly better in segmenting small classes, \textcolor{rebuttal}{even in the presence of occlusions.} \textcolor{rebuttal}{Specifically, in case (a), SegFormer achieves more detailed segmentation in the yellow cystic duct’s top-left and bottom-right regions, which is crucial for successful surgeries. Additionally, in case (b), using the Mask2Former model, our pipeline significantly enhances the segmentation of tiny, barely visible blood clots. In case (c), where slight smoke is present, our method is particularly effective, accurately segmenting the second blue hepatic vein area at the bottom of the image where other approaches failed. Moreover, in case (d), in a highly smoke-obscured scenario, our result closely aligns with the ground truth, especially in capturing the contours of blood, demonstrating the robustness of our approach in maintaining critical detail even under severe visual impairment.}

The success \textcolor{rebuttal}{in addressing underrepresented classes} lies in CASDM's enhanced diversity and realism, improving textural and structural details vital for accurate diagnostics. The integration of synthesized segmentation maps further enables the model to output images with diverse structures, increasing alignment with unseen testing data. This pipeline enriches the training set for downstream segmentation models with novel patterns, significantly boosting their generalization capabilities \textcolor{rebuttal}{for underrepresented classes}.

\textcolor{rebuttal}{In our evaluation of well-represented classes using several data expansion methods, as shown in Table \ref{table3}, we observed less significant improvements. For the neither scarce, nor small connective tissue class, adding synthetic data had minimal impact on the segmentation performance. In the case of the rare, but highly uniform and large liver ligament class, the best result came from CASDM alone, with a 0.9\% increase in mIoU and a 0.5\% increase in mDice on SegFormer compared to the original dataset. For the common and moderately sized L-hook electrocautery class, CASDM on Mask2Former yielded a 0.8\% increase in mIoU and a 0.4\% increase in mDice. These results suggest that the downstream segmentation models have already learned sufficient features of these classes from the original dataset, making additional data less impactful. Improvements in underrepresented classes, such as the cystic duct, could even reach up to 3.2\% in mIoU and
1.9\% in mDice. The results for CASDM + synthetic maps were generally comparable to using CASDM alone, which is expected since generating more diverse maps for classes with consistent shapes and locations is not meaningful. Furthermore, when the original segmentation maps already contain ample examples with almost all possible shapes and locations, adding more maps offers diminishing returns. These findings highlight that our model is best suited for balancing imbalanced datasets, particularly for small and scarce classes. Over-representing well-represented classes can lead to unnecessary dominance, which ultimately reinforces our primary goal of addressing data imbalance by focusing on underrepresented classes.}

\begin{table*}[t]
\centering
\caption{\textcolor{rebuttal}{Segmentation performance of our methods on the gastrectomy SISVSE dataset.}}
\textcolor{rebuttal}{\resizebox{\textwidth}{!} {
\begin{tabular}{|l|l|c|c|c|c|c|c|c|c|}
  \hline
  \multirow{2}{*}{\textbf{Seg. Model}} & \multirow{2}{*}{\textbf{Training Dataset}} & \multicolumn{2}{c|}{\textbf{Suction Irrigation}} & \multicolumn{2}{c|}{\textbf{Endotip}} & \multicolumn{2}{c|}{\textbf{Spleen}} & \multicolumn{2}{c|}{\textbf{All}} \\\cline{3-10}
  &  & \textbf{mIoU} $\uparrow$ & \textbf{mDice} $\uparrow$ & \textbf{mIoU} $\uparrow$ & \textbf{mDice} $\uparrow$ & \textbf{mIoU} $\uparrow$ & \textbf{mDice} $\uparrow$ & \textbf{mIoU} $\uparrow$ & \textbf{mDice} $\uparrow$ \\\hline
  \multirow{3}{*}{SegFormer} & Original & 73.1 & 84.4 & 87.9 & 93.6 & 39.2 & 56.3 & 68.0 & 79.9 \\\cline{2-10}
  & + CASDM & 74.5 & 85.4 & 86.7 & 92.9 & 45.0 & 62.0 & 68.6 & 80.4 \\\cline{2-10}
  & + CASDM + Synthetic Maps & \textbf{76.4} & \textbf{86.6} & \textbf{88.9} & \textbf{94.1} & \textbf{49.7} & \textbf{66.4} & 69.3 & 80.9 \\\hline
  \multirow{3}{*}{Mask2Former} & Original & 74.6 & 85.4 & 83.0 & 90.7 & 35.0 & 51.9 & 64.3 & 77.4 \\\cline{2-10}
  & + CASDM & 75.4 & 86.0 & 84.7 & 91.7 & 38.9 & 56.0 & 67.2 & 79.6 \\\cline{2-10}
  & + CASDM + Synthetic Maps & 75.5 & 86.0 & 87.7 & 93.4 & 44.8 & 61.9 & \textbf{69.4} & \textbf{81.3} \\\hline
\end{tabular}}
}
\label{table4}
\end{table*}

\begin{table}[H]
\centering
\caption{\textcolor{rebuttal}{Ablation study on model components.}}
\resizebox{8.5cm}{!} { 
\begin{tabular}{|l|c|c|c|}
  \hline
  \textbf{Method} & \textcolor{rebuttal}{\textbf{FID $\downarrow$}} & \textbf{Mean MS-SSIM $\uparrow$} & \textbf{Mean PSNR $\uparrow$} \\\hline
  SDM & \textcolor{rebuttal}{85.51} & 0.60 & 13.08 \\\hline
  SDM + \( \mathcal{L}_{\text{CAMSE}} \) & \textcolor{rebuttal}{84.62} & \textbf{0.64} & 14.32 \\\hline
  SDM + \( \mathcal{L}_{\text{CASP}} \) & \textcolor{rebuttal}{81.59} & 0.62 & 13.38 \\\hline
  CASDM & \textcolor{rebuttal}{\textbf{75.89}} & \textbf{0.64} & \textbf{14.44} \\\hline
\end{tabular}
}
\label{table5}
\end{table}

\emph{\textbf{Ablation Study on Model Components: }}The ablation study in Table \ref{table5} shows that the addition of both \( \mathcal{L}_{\text{CAMSE}} \) and \( \mathcal{L}_{\text{CASP}} \) respectively improves the quality of synthesised images in terms of \textcolor{rebuttal}{FID}, Mean MS-SSIM and Mean PSNR. When combined in our CASDM, they achieve the \textcolor{rebuttal}{best} performance, indicating the efficiency of both modules.

\textcolor{rebuttal}{\emph{\textbf{Generalisability Across Datasets: }} We extended our experiments to include another surgical procedure, gastrectomy, using the publicly available SISVSE dataset \cite{yoon2022surgical}. Our pipeline was retrained on this dataset, focusing on three less-represented classes: Suction Irrigation, Endotip, and Spleen. These classes appear in 8.36\%, 8.76\%, and 9.29\% of the images, respectively, and occupy 3.85\%, 4.57\%, and 2.47\% of the image area in those images. All settings were kept consistent with those used on the cholecystectomy dataset. Using the original and augmented gastrectomy datasets, we trained and tested the SegFormer and Mask2Former models. The results, detailed in Table \ref{table4}, show an average mIoU improvement of 3.6\% and mDice improvement of 4.3\% for these three classes on SegFormer when using our full pipeline, which includes both segmentation map generation and image generation. On Mask2Former, the mIoU and mDice for these three classes improved by 5.1\% and 4.4\%, respectively. Our pipeline achieved more significant improvements on the gastrectomy dataset, due to dataset's greater diversity and imbalance, with 32 classes where the selected classes appeared in a smaller percentage of images and occupied a smaller area. This outcome demonstrates the strong generalisability of our pipeline, particularly in more challenging and diverse datasets.}

\section{Conclusions}

In this work, we proposed CASDM, a novel image synthesis approach designed to address data scarcity and imbalance. We also introduced a pipeline that generates segmentation maps and corresponding images from text prompts. CASDM leverages class-aware MSE and self-perceptual losses to create diverse, high-quality images that closely adhere to segmentation maps, enhancing the representation of less visible classes. Our evaluation confirms that this approach significantly improves both the quality of synthesized images and the performance of segmentation models, demonstrating its potential to enhance surgical outcomes by providing robust and balanced training data, particularly for endoscopic procedures. \textcolor{rebuttal}{ Additionally, our results highlight the strong generalisability of our pipeline across different surgical procedures, showcasing its broader applicability.}

\textcolor{rebuttal}{While our method is not real-time, the fact that data augmentation is performed offline makes real-time inference not a concern for applications. Our next step is to develop a model capable of directly generating segmentation maps and closely aligned images from text prompts in a single step, further enhancing applicability and ultimately supporting better surgical decision-making and outcomes.}

\section{\textcolor{rebuttal}{Acknowledgement}}

\textcolor{rebuttal}{Dr. Giannarou and Mr. Zhou have been supported by the Royal Society [URF$\setminus$R$\setminus$201014].}

\bibliographystyle{unsrt}
\renewcommand{\refname}{} 
\section{References}  

\bibliography{reference}

\begin{thebibliography}{10}

\bibitem{madani2022artificial}
Amin Madani, Babak Namazi, Maria~S Altieri, Daniel~A Hashimoto, Angela~Maria Rivera, Philip~H Pucher, Allison Navarrete-Welton, Ganesh Sankaranarayanan, L~Michael Brunt, Allan Okrainec, et~al.
\newblock Artificial intelligence for intraoperative guidance: using semantic segmentation to identify surgical anatomy during laparoscopic cholecystectomy.
\newblock {\em Annals of surgery}, 276(2):363--369, 2022.

\bibitem{ali2021deep}
Sharib Ali, Mariia Dmitrieva, Noha Ghatwary, Sophia Bano, Gorkem Polat, Alptekin Temizel, Adrian Krenzer, Amar Hekalo, Yun~Bo Guo, Bogdan Matuszewski, et~al.
\newblock Deep learning for detection and segmentation of artefact and disease instances in gastrointestinal endoscopy.
\newblock {\em Medical image analysis}, 70:102002, 2021.

\bibitem{kolbinger2023anatomy}
Fiona~R Kolbinger, Franziska~M Rinner, Alexander~C Jenke, Matthias Carstens, Stefanie Krell, Stefan Leger, Marius Distler, J{\"u}rgen Weitz, Stefanie Speidel, and Sebastian Bodenstedt.
\newblock Anatomy segmentation in laparoscopic surgery: comparison of machine learning and human expertise--an experimental study.
\newblock {\em International Journal of Surgery}, 109(10):2962--2974, 2023.

\bibitem{chen2017rethinkingatrousconvolutionsemantic}
Liang-Chieh Chen, George Papandreou, Florian Schroff, and Hartwig Adam.
\newblock Rethinking atrous convolution for semantic image segmentation, 2017.

\bibitem{ronneberger2015unetconvolutionalnetworksbiomedical}
Olaf Ronneberger, Philipp Fischer, and Thomas Brox.
\newblock U-net: Convolutional networks for biomedical image segmentation, 2015.

\bibitem{xie2021segformersimpleefficientdesign}
Enze Xie, Wenhai Wang, Zhiding Yu, Anima Anandkumar, Jose~M. Alvarez, and Ping Luo.
\newblock Segformer: Simple and efficient design for semantic segmentation with transformers, 2021.

\bibitem{cheng2022maskedattentionmasktransformeruniversal}
Bowen Cheng, Ishan Misra, Alexander~G. Schwing, Alexander Kirillov, and Rohit Girdhar.
\newblock Masked-attention mask transformer for universal image segmentation, 2022.

\bibitem{psychogyios2024sarrarp50segmentationsurgicalinstrumentation}
Dimitrios Psychogyios, Emanuele Colleoni, Beatrice~Van Amsterdam, Chih-Yang Li, Shu-Yu Huang, Yuchong Li, Fucang Jia, Baosheng Zou, Guotai Wang, Yang Liu, Maxence Boels, Jiayu Huo, Rachel Sparks, Prokar Dasgupta, Alejandro Granados, Sebastien Ourselin, Mengya Xu, An~Wang, Yanan Wu, Long Bai, Hongliang Ren, Atsushi Yamada, Yuriko Harai, Yuto Ishikawa, Kazuyuki Hayashi, Jente Simoens, Pieter DeBacker, Francesco Cisternino, Gabriele Furnari, Alex Mottrie, Federica Ferraguti, Satoshi Kondo, Satoshi Kasai, Kousuke Hirasawa, Soohee Kim, Seung~Hyun Lee, Kyu~Eun Lee, Hyoun-Joong Kong, Kui Fu, Chao Li, Shan An, Stefanie Krell, Sebastian Bodenstedt, Nicolas Ayobi, Alejandra Perez, Santiago Rodriguez, Juanita Puentes, Pablo Arbelaez, Omid Mohareri, and Danail Stoyanov.
\newblock Sar-rarp50: Segmentation of surgical instrumentation and action recognition on robot-assisted radical prostatectomy challenge, 2024.

\bibitem{ayobi2023matis}
Nicol{\'a}s Ayobi, Alejandra P{\'e}rez-Rond{\'o}n, Santiago Rodr{\'\i}guez, and Pablo Arbel{\'a}ez.
\newblock Matis: Masked-attention transformers for surgical instrument segmentation.
\newblock In {\em 2023 IEEE 20th International Symposium on Biomedical Imaging (ISBI)}, pages 1--5. IEEE, 2023.

\bibitem{10.1145/3422622}
Ian Goodfellow, Jean Pouget-Abadie, Mehdi Mirza, Bing Xu, David Warde-Farley, Sherjil Ozair, Aaron Courville, and Yoshua Bengio.
\newblock Generative adversarial networks.
\newblock {\em Commun. ACM}, 63(11):139–144, oct 2020.

\bibitem{mahapatra2018efficient}
Dwarikanath Mahapatra, Behzad Bozorgtabar, Jean-Philippe Thiran, and Mauricio Reyes.
\newblock Efficient active learning for image classification and segmentation using a sample selection and conditional generative adversarial network.
\newblock In {\em International Conference on Medical Image Computing and Computer-Assisted Intervention}, pages 580--588. Springer, 2018.

\bibitem{thanh2020catastrophic}
Hoang Thanh-Tung and Truyen Tran.
\newblock Catastrophic forgetting and mode collapse in gans.
\newblock In {\em 2020 international joint conference on neural networks (ijcnn)}, pages 1--10. IEEE, 2020.

\bibitem{DBLP:journals/corr/abs-2006-11239}
Jonathan Ho, Ajay Jain, and Pieter Abbeel.
\newblock Denoising diffusion probabilistic models.
\newblock {\em CoRR}, abs/2006.11239, 2020.

\bibitem{DBLP:journals/corr/abs-2010-02502}
Jiaming Song, Chenlin Meng, and Stefano Ermon.
\newblock Denoising diffusion implicit models.
\newblock {\em CoRR}, abs/2010.02502, 2020.

\bibitem{ozbey2023unsupervised}
Muzaffer {\"O}zbey, Onat Dalmaz, Salman~UH Dar, Hasan~A Bedel, {\c{S}}aban {\"O}zturk, Alper G{\"u}ng{\"o}r, and Tolga {\c{C}}ukur.
\newblock Unsupervised medical image translation with adversarial diffusion models.
\newblock {\em IEEE Transactions on Medical Imaging}, 2023.

\bibitem{kazerouni2023diffusion}
Amirhossein Kazerouni, Ehsan~Khodapanah Aghdam, Moein Heidari, Reza Azad, Mohsen Fayyaz, Ilker Hacihaliloglu, and Dorit Merhof.
\newblock Diffusion models in medical imaging: A comprehensive survey.
\newblock {\em Medical Image Analysis}, page 102846, 2023.

\bibitem{wang2022semantic}
Weilun Wang, Jianmin Bao, Wengang Zhou, Dongdong Chen, Dong Chen, Lu~Yuan, and Houqiang Li.
\newblock Semantic image synthesis via diffusion models.
\newblock {\em arXiv preprint arXiv:2207.00050}, 2022.

\bibitem{du2023arsdm}
Yuhao Du, Yuncheng Jiang, Shuangyi Tan, Xusheng Wu, Qi~Dou, Zhen Li, Guanbin Li, and Xiang Wan.
\newblock Arsdm: colonoscopy images synthesis with adaptive refinement semantic diffusion models.
\newblock In {\em International conference on medical image computing and computer-assisted intervention}, pages 339--349. Springer, 2023.

\bibitem{zhuang2023semantic}
Yan Zhuang, Benjamin Hou, Tejas~Sudharshan Mathai, Pritam Mukherjee, Boah Kim, and Ronald~M Summers.
\newblock Semantic image synthesis for abdominal ct.
\newblock In {\em International Conference on Medical Image Computing and Computer-Assisted Intervention}, pages 214--224. Springer, 2023.

\bibitem{twinanda2016endonetdeeparchitecturerecognition}
Andru~P. Twinanda, Sherif Shehata, Didier Mutter, Jacques Marescaux, Michel de~Mathelin, and Nicolas Padoy.
\newblock Endonet: A deep architecture for recognition tasks on laparoscopic videos, 2016.

\bibitem{yoon2022surgical}
Jihun Yoon, SeulGi Hong, Seungbum Hong, Jiwon Lee, Soyeon Shin, Bokyung Park, Nakjun Sung, Hayeong Yu, Sungjae Kim, SungHyun Park, et~al.
\newblock Surgical scene segmentation using semantic image synthesis with a virtual surgery environment.
\newblock In {\em International Conference on Medical Image Computing and Computer-Assisted Intervention}, pages 551--561. Springer, 2022.

\bibitem{ho2022classifier}
Jonathan Ho and Tim Salimans.
\newblock Classifier-free diffusion guidance.
\newblock {\em arXiv preprint arXiv:2207.12598}, 2022.

\bibitem{lin2024diffusionmodelperceptualloss}
Shanchuan Lin and Xiao Yang.
\newblock Diffusion model with perceptual loss, 2024.

\bibitem{radford2021learningtransferablevisualmodels}
Alec Radford, Jong~Wook Kim, Chris Hallacy, Aditya Ramesh, Gabriel Goh, Sandhini Agarwal, Girish Sastry, Amanda Askell, Pamela Mishkin, Jack Clark, Gretchen Krueger, and Ilya Sutskever.
\newblock Learning transferable visual models from natural language supervision, 2021.

\bibitem{lu2023dpmsolverfastsolverguided}
Cheng Lu, Yuhao Zhou, Fan Bao, Jianfei Chen, Chongxuan Li, and Jun Zhu.
\newblock Dpm-solver++: Fast solver for guided sampling of diffusion probabilistic models, 2023.

\bibitem{woodland2024featureextractiongenerativemedical}
McKell Woodland, Austin Castelo, Mais~Al Taie, Jessica Albuquerque~Marques Silva, Mohamed Eltaher, Frank Mohn, Alexander Shieh, Suprateek Kundu, Joshua~P. Yung, Ankit~B. Patel, and Kristy~K. Brock.
\newblock Feature extraction for generative medical imaging evaluation: New evidence against an evolving trend, 2024.

\end{thebibliography}

\end{document}